\documentclass[smallextended]{svjour3}

\smartqed  
\usepackage{booktabs}
\usepackage{enumitem}
\usepackage{longtable}
\usepackage{graphicx}
\usepackage{array}
\usepackage{dcolumn}
\usepackage{amsfonts}
\usepackage{amssymb}
\usepackage{xspace}
\usepackage{xfrac}
\usepackage{array}
\usepackage{multirow}
\usepackage{xcolor}
\usepackage{footnote}
\usepackage{url}
\usepackage{makeidx}
\usepackage{epsfig}
\usepackage{amsmath}
\usepackage{multicol}
\usepackage{mathtools}
\usepackage[misc]{ifsym}
\usepackage[ruled,linesnumbered]{algorithm2e}
\usepackage{tabularx}
\usepackage{array}

\newcolumntype{C}[1]{>{\centering\arraybackslash}m{#1}}

\usepackage{caption}
\usepackage{subcaption}
\begin{document}

\title{\texttt{VEST}: Automatic Feature Engineering for Forecasting}

\titlerunning{Automatic Feature Engineering for Forecasting}

\author{Vitor Cerqueira, Nuno Moniz and Carlos~Soares}

\authorrunning{V. Cerqueira et al.}

\institute{Vitor Cerqueira (\Letter) \at
         INESC TEC, Porto, Portugal\\
         \email{cerqueira.vitormanuel@gmail.com}
         \and
         Nuno Moniz \at
         INESC TEC, Porto, Portugal\\
         University of Porto, Porto, Portugal\\
         \email{nmmoniz@inesctec.pt}
         \and
         Carlos Soares \at
         Fraunhofer AICOS Portugal, Porto, Portugal\\
         INESC TEC, Porto, Portugal\\
         University of Porto, Porto, Portugal\\
         \email{csoares@fe.up.pt}
         }

\date{Received: date / Accepted: date}

\maketitle

\begin{abstract}

Time series forecasting is a challenging task with applications in a wide range of domains. 
Auto-regression is one of the most common approaches to address these problems. Accordingly, observations are modelled by multiple regression using their past lags as predictor variables.
We investigate the extension of auto-regressive processes using statistics which summarise the recent past dynamics of time series. 
The result of our research is a novel framework called \texttt{VEST}, designed to perform feature engineering using univariate and numeric time series automatically.
The proposed approach works in three main steps. First, recent observations are mapped onto different representations. 
Second, each representation is summarised by statistical functions. 
Finally, a filter is applied for feature selection. 
We discovered that combining the features generated by \texttt{VEST} with auto-regression significantly improves forecasting performance. We provide evidence using 90 time series with high sampling frequency. \texttt{VEST} is publicly available online.

\end{abstract}

\section{Introduction} 

Analysing and learning from time series is one of the most active topics in scientific research. One of the tasks related to this topic is forecasting, which denotes the process of predicting the value of future observations given some historical data. This task is relevant to organisations across a wide range of domains of application. In many of these organisations, forecasting processes can have a significant financial impact \cite{kahn2003measure}.

In the machine learning literature, it is widely accepted that the feature set used to represent a data set is a crucial component for building accurate predictive models \cite{guyon2006introduction}. Hence, feature engineering is regarded as a critical step in machine learning projects. 
However, feature engineering is often an \textit{ad-hoc} process. Practitioners design new features based on their domain knowledge and expertise.
The limitations of current approaches to feature engineering are particularly relevant in time series forecasting, where, although evidence exists that it improves forecasting performance \cite{Oliveira2014EnsemblesFT}, it is often overlooked.

Time series forecasting tasks are typically formalised using an auto-regressive approach. Accordingly, observations are modelled using multiple regression; the future observations we want to predict represent the target variable, and the past lags of these observations are used as explanatory variables. Auto-regression is at the core of many forecasting models in the literature, such as, for example, ARIMA (Auto-Regressive Integrated Moving Average) \cite{box2015time}.
This approach presents an opportunity to approach feature engineering in a systematic way. 
Statistical features can be extracted from recent observations of time series. These features can help summarise the dynamics of the data and enrich its representation. For example, a simple feature such as the average of the past recent observations can help capture the overall level of the time series in a given point in time. 
If the process is adequately done, new features will lead to more accurate predictive models and better forecasting performance.

\subsection{Our Approach and Paper Organisation}

Despite having domain expertise, professionals often lack proper time series analysis skills \cite{taylor2018forecasting}. Getting the most out of state of the art time series forecasting methods requires significant experience, and it is a complex and time-consuming task. In this context, developing a framework for automatically extracting an optimal representation from an input time series is beneficial for practitioners with minimal technical expertise.

This paper presents and describes an automatic feature engineering approach called \texttt{VEST} (\textbf{Ve}ctor of \textbf{S}tatistics from \textbf{T}ime series). 
\texttt{VEST} is specifically designed to address forecasting problems. 
To the best of our knowledge, this is the first approach to automatically and systematically extract features from time series.

\texttt{VEST} works according to the following three main steps:

\begin{itemize}
    \item Transformation: We transform the time series into several distinct representations. This process may be beneficial for describing from different perspectives the underlying process causing the time series. For example, a simple moving average transformation can be useful to remove the spurious behaviour of time series;
    
    \item Summarisation: Each representation is summarised using statistics (e.g. mean, standard deviation);
    
    \item Selection: The first two steps can lead to a high dimensional problem. We apply a feature selection procedure to cope with this issue. The final set of features is concatenated with the original representation (before any transformation) to learn a regression model for forecasting.

\end{itemize}

In this paper, we show significant performance gains when applying \texttt{VEST}, based on a case study comprised of 90 univariate time series from several domains of application. These improvements are evidenced for predicting the next point of the time series using two different learning algorithms: a variant of the model tree \cite{quinlan1993combining}, and lasso  \cite{tibshirani1996regression}.

\texttt{VEST} is available online at \url{https://github.com/vcerqueira/vest}. Additionally, the code necessary to reproduce the experimental evaluation presented in the paper is made available to encourage reproducible research.

The organisation of the paper is as follows. In Section \ref{sec:pd}, we formalise the time series forecasting problems as a predictive regression task. We present a literature review in Section~\ref{sec:rw}, including topics related to the feature engineering process and its automation, feature selection, and time series representation and feature extraction. Section~\ref{sec:feng} presents \texttt{VEST}, formalising its main steps: transforming the original representation, summarisation using statistics, and feature selection. We show the usefulness of this framework using empirical evidence presented in Section \ref{sec:exp}. We present a discussion on the results from such experiments, challenges, and future directions in Section \ref{sec:disc}. The paper is concluded in Section~\ref{sec:conclusions}.

\section{Problem Definition}\label{sec:pd}

A univariate time series represents a temporal sequence of values $Y = \{y_1, y_2, \dots,$ $y_n \}$, where $y_i \in \mathcal{Y} \subset \mathbb{R}$ is the value of $Y$ at time $i$ and $n$ is the length of $Y$. Forecasting denotes the process of predicting the value of the upcoming observations of the time series, $y_{n+1}, \ldots, y_{n+h}$, 
given the historical past observations, where $h$ denotes the forecasting horizon. In this work, we focus on $h = 1$, which means we attempt to forecast the next value of the time series.
We adopt an auto-regressive approach to address the problem of time series forecasting. Accordingly, observations of a time series are regressed on their past lags. 

We start by reconstructing the time series as a geometric object by applying a time delay embedding using the Takens theorem~\cite{Takens1981}. 
Then, the predictive task is framed as a multiple regression problem.
We construct a set of observations of the form ($X$, $y$). In each observation, the value of $y_i$ is modelled based on the past $p$ values before it: $X_i = \{y_{i-1}, y_{i-2}, \dots, y_{i-p} \}$, where $y_i \in \mathcal{Y} \subset \mathbb{R}$,
which represents the observation we want to predict, and $X_i \in \mathcal{X} \subset \mathbb{R}^p$ represents the $i$-th embedding vector. Effectively, the time series is transformed into the data set $\mathcal{D}(X,y) = \{X_i, y_i\}^{n}_{p+1}$.

The learning objective is to build a regression model that provides an approximation to an unknown function $f : \mathcal{X} \rightarrow \mathcal{Y}$.
The principle behind this approach is to model the conditional distribution of the $i$-th value of the time series given its $p$ past values: $f$($y_{i} | X_i$). 

\section{Related Research}\label{sec:rw}

This section provides an overview of the literature related to the topic of this work. First, we describe the importance of feature engineering and outline automatic procedures to address this task (Section~\ref{sec:rw_fe}). 
Afterwards, we focus on time series data. We describe approaches for changing the representation of time series (Section~\ref{sec:reps}). Finally, we overview approaches for extracting features from time series with the objective of predictive modelling (Section~\ref{sec:tsfex}).  

\subsection{Feature Engineering}\label{sec:rw_fe}

The goal of feature engineering is to enrich the representation of a data set with additional explanatory variables. The expectation is that such new predictors contain useful information and lead to more accurate predictive models~\cite{guyon2006introduction}. 

In order to clarify the scope of our work, we split the generated variables through feature engineering into two classes: exogenous and endogenous.
\emph{Exogenous variables} are those derived from an external source.
Consider an example of a time series representing the number of rooms occupied per day in a hotel. 
Forecasting the values of such a time series is interesting to the business for different reasons (e.g. pricing).
In this scenario, a simple binary variable containing the information regarding whether or not the observation to be predicted occurs during the weekend may be useful to the predictive model.
Since information is not contained within the original observations (each $y \in \mathcal{Y}$) we call it exogenous.

Regarding \emph{endogenous variables}, $X_i$ represents the embedding vector (Section \ref{sec:pd}) of a time series in a given point in time $i$. We can try to derive more information from $X_i$ by applying some transformations or summary statistics. For example, the average of the values of $X_i$ in a specific period may be a useful indicator for describing the overall level of the time series at that point. As such, a new explanatory variable is generated based on the time series itself, i.e. an endogeneous feature.

In this work, we focus on the automatic discovery of endogenous features in numeric time series. The goal is to augment the representation of the embedding vectors and improve forecasting performance of predictive models.

\subsubsection{Automatic Feature Engineering}

Typically, feature engineering is an \textit{ad-hoc} process~\cite{guyon2006introduction,pinto2016towards}. Practitioners design features based on their domain knowledge and expertise. However, this process is time-consuming, and requires both domain expertise and imagination.\footnote{\url{https://www.kdnuggets.com/2019/03/why-automl-wont-replace-data-scientists.html}}

Recent approaches have been developed to systematise the feature engineering process. This research line is designated in the literature as automatic feature engineering.
Examples include the following: Deep Feature Synthesis~\cite{kanter2015deep}, ExploreKit~\cite{katz2016explorekit}, AutoLearn~\cite{kaul2017autolearn}, Cognito~\cite{khurana2016cognito}, or OneBM~\cite{lam2017one}. These frameworks focus on discovering relevant features from data sets comprised of several attributes, which can be either categorical or numeric. Moreover, most of these focus solely on classification problems and i.i.d. data. In this work, we focus on univariate time series data and the problem of forecasting. As such, many of the operations we develop to discover new relevant features intend to leverage the temporal dependency among the observations in the data set.

Deep learning methods can also be regarded as having an internal automatic feature engineering component. These approaches are able to learn higher-order representations based on the raw input data. Still, there are important factors which make standard feature engineering relevant.
Deep learning models require a large amount of data, which is often not readily available. 
The internal representations of neural networks are opaque, while standard feature engineering is typically based on interpretable operations. This transparency may be important in sensitive applications. 
Besides, the two approaches are not incompatible, as neural networks can potentially leverage standard feature engineering, for example, to improve their learning efficiency.

\subsection{Time Series Representation}\label{sec:reps}

Sometimes, time series are analysed using a representation that is different from the original one. Changing the representation of a time series can be beneficial for (i) reducing the dimensionality of the data, which leads to more efficient storage and processing; (ii) implicit handling of noise; and (iii) focusing on fundamental characteristics of the data~\cite{esling2012time}. 
We refer to the work by Esling and Agon~\cite{esling2012time} for a complete read on time series transformations.

Keogh~\cite{keogh2004towards} splits time series representation methods into three main types: non-data adaptive, data-adaptive, and model-based. In non-data adaptive approaches, the parameters of the transformation are independent of the underlying data. Examples of this approach are the discrete wavelet transformation~\cite{percival2006wavelet} or the simple moving average. 
Conversely, the parameters of data-adaptive methods depend to some extent on the time series. Symbolic aggregate approximation~\cite{lin2003symbolic} is a well-known approach of this sort. 

Model-based approaches work on the assumption that some underlying model generates the time series. As such, parameters of the model represent the time series. Auto-regressive moving average (ARMA)~\cite{chatfield2000time} models are an example of this type.


\subsection{Time Series Feature Extraction}\label{sec:tsfex}

Extracting features from time series has been shown to improve performance in different tasks such as forecasting and classification~\cite{prudencio2004using,christ2016distributed}.
We split the literature on this topic into two dimensions: sequence descriptions and sub-sequence descriptions. The former denotes approaches which summarise the complete set of observations available in a time series. The latter extract features from sub-sequences of time series, i.e. the embedding vectors.

\subsubsection{Sequence descriptions}

There are several approaches which extract features from the complete time series to improve forecasting performance. Examples of this are the works of Prudêncio and Ludermir~\cite{prudencio2004using}, Lemke et al.~\cite{lemke2010meta}, Barandas et al.~\cite{barandas2020tsfel}, or Kang et al.~\cite{kang2017visualising}.
Often, the goal of these approaches is to use meta-learning for model selection or combination. Essentially, they extract features from each time series in a given database. Then, a predictive model is created which relates the features of a time series with the most appropriate forecasting model in that data. In effect, for a new given time series, such a meta-learning model can make predictions regarding which model, or set of models, is more appropriate.
Recently, Montero-Manso et al.~\cite{montero2020fforma} applied this type of approach and ranked second in the well-known forecasting M4-competition.

In the context of time series classification, Christ et al.~\cite{christ2016distributed} proposed the method FRESH for feature engineering. This method automatically extracts a large number of features from each time series in the database and selects the most relevant ones for building the classifier. 
Fulcher et al. \cite{fulcher2017hctsa} presented a feature extraction framework called \textit{hctsa} for time series analysis. This tool extracts over 7000 features.
Lubba et al. \cite{lubba2019catch22} selected a a subset of 22 \textit{hctsa} features, leading to a 1000-fold reduction in computation time for feature extraction and only a small reduction in time series classification performance.

\subsubsection{Sub-sequence descriptions}

Compared to feature extraction from complete time series, few works are leveraging these processes for time series sub-sequences. 
Paras et al. \cite{paras2009feature} used a set of statistics, such as simple and exponential moving averages, to improve a neural network model for weather forecasting. 
Oliveira and Torgo \cite{Oliveira2014EnsemblesFT} show that the average and standard deviation of recent values improve the performance of a bagging ensemble. Cerqueira et al. \cite{cerqueira2017dets} later corroborated these results using heterogeneous ensembles.
However, the potential of the systematic application of approaches deriving new features from sub-sequences of time series has never been explored.

We follow this research line and derive new features from sub-sequences of time series. To accomplish this, we develop \texttt{VEST}, a novel framework for automatic feature engineering using univariate time series. \texttt{VEST} extracts new features from embedding vectors representing a time series and selects the most important ones for combination with the original vector.

\section{\texttt{VEST}: Vector of Statistics from Time series}\label{sec:feng}

In this section, we propose and formalise \texttt{VEST} (for \textbf{Ve}ctor of \textbf{S}tatistics from \textbf{T}ime series), an automatic feature engineering process for univariate and numeric time series. \texttt{VEST} is specifically designed for forecasting problems. Given a time series $Y = \{y_1, \dots, y_n\}$, the goal is to predict the value of the next observation, $y_{n+1}$. 
Following the formalisation presented in Section \ref{sec:pd}, we address time series forecasting as an auto-regressive task. The $i$-th observation of a time series, $y_i$, is modelled according to the $i$-th embedding vector $X_i = \{y_{i-1}, y_{i-2},\dots, y_{i-p}\}$, which represents the $p$ previous observations.

\texttt{VEST} is based on the manipulation of the embedding vectors representing each observation of a time series. Particularly, the method contains three main steps, addressing feature generation (steps 1 and 2) and selection (3):

\begin{enumerate}
    \item Transforming each embedding vector $X$ into different representations (Section~\ref{subsubsec:transform});
    \item Summarising each representation into features using statistical functions (Section~\ref{subsubsec:summary});
    \item Selection of relevant features (Section~\ref{subsec:selection}).
\end{enumerate}

\noindent We adopt an expand--reduce approach for feature engineering~\cite{do2019automated}. In the first two steps of the methodology, we expand the representation of the data with a large set of features. In the final step, we reduce this representation and keep only the most relevant variables.

\begin{figure}[hbt]
\centering
\includegraphics[width=.8\textwidth]{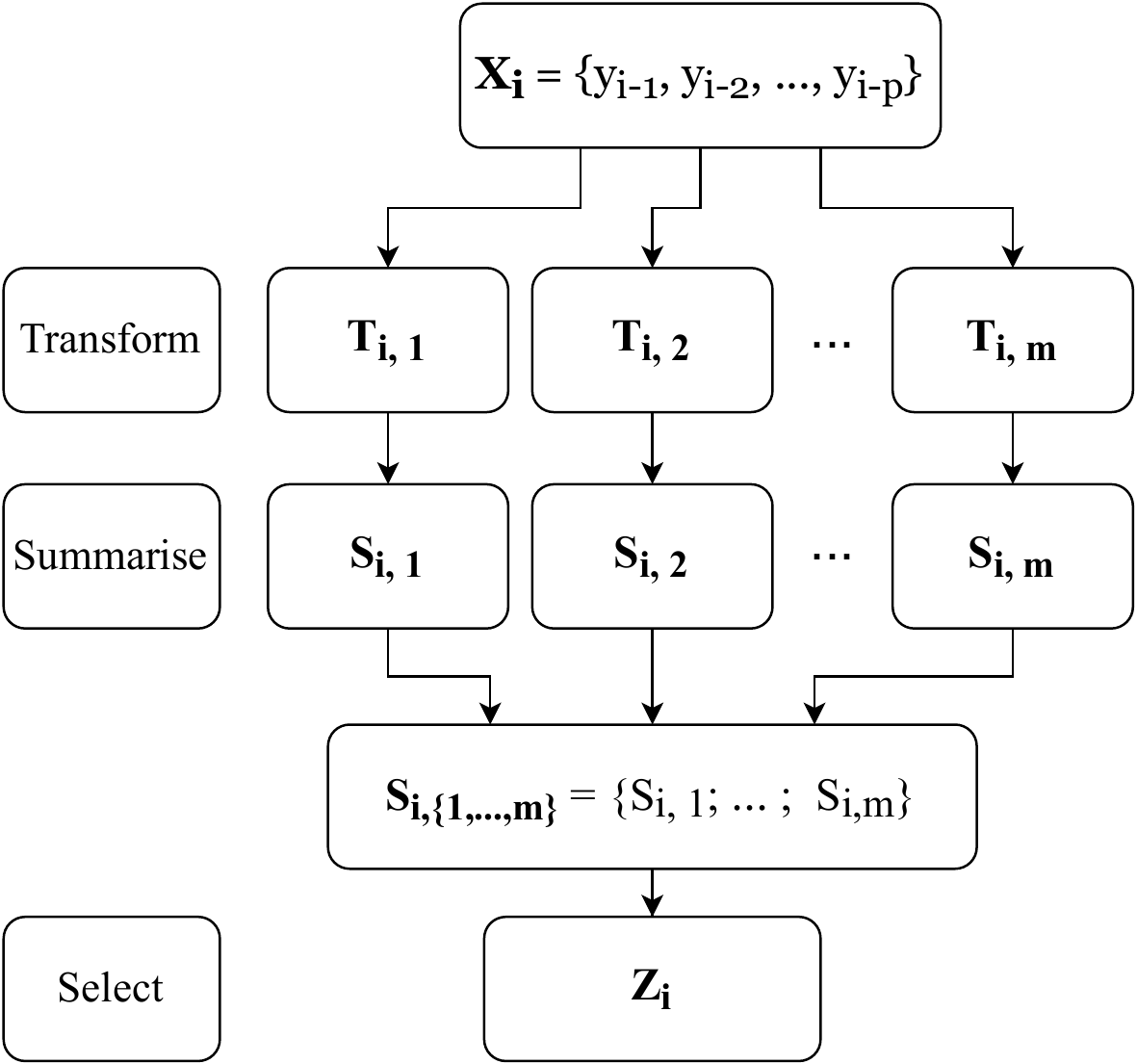}
\caption{Feature engineering workflow for a given embedding vector $X_i$. $X_i$ is mapped into $m$ different representations, $\{T_{i,1}, \dots, T_{i,m} \}$. Each representation is summarised into a set of statistics $S_{i,m}$. For example, $S_{i,1}$ denotes a set of features constructed from the representation $T_{i,1}$. Finally, feature selection is carried out, leading to the final set of features $Z_i$.}
\label{fig:fescheme}
\end{figure}

In the next subsections, we formalise these steps in more detail. The workflow for a given instance $X_i$ is exemplified in Figure \ref{fig:fescheme}.

\subsection{Feature Generation}

We base our feature generation process on the manipulation of each embedding vector $X \in \mathcal{X}$. In this sense, our approach is entirely endogenous. Exogenous features (e.g. holiday information) can also be essential to improve forecasting performance. However, such analysis is out of the scope of this work.

\subsubsection{Transform Operations}\label{subsubsec:transform}

The first step of \texttt{VEST} is a transformation procedure. This procedure generates new representations from the original embedding vectors $X$. As we mentioned in Section \ref{sec:reps}, changing the representation of time series embedding vectors is beneficial for handling noise, and to focus on essential characteristics of the data \cite{esling2012time}. We hypothesize that different representations obtained by distinct transformation operations generate new, complementary information regarding the dynamics of the time series. 
Therefore, this combination of these new types of information can lead to improvements in forecasting performance which cannot be obtained by using each one of them separately. 
Formally, a transform operation maps an embedding vector $X_i$ into another $q$-dimensional vector $T$:

\begin{align*}
  \text{Transform}_j: \mathbb{R}^p &\to \mathbb{R}^q\\
  X_i &\mapsto T_{i,j}
\end{align*}

\noindent Essentially, $X_i$ is mapped onto $T_{i,j}$, $\forall$ $j \in \{1, \dots, m\}$. Hence, $T_{i,j}$ is a vector which denotes the $j$-th representation of the $i$-th embedding vector.

An example of a possible transform operation is differencing, which means the differences between consecutive observations. This transformation is often applied to time series to remove the trend component. 

\subsubsection{Summary Operations}\label{subsubsec:summary}

By applying distinct transform operations, an embedding vector has several representations (one for each such operation). The next step in the methodology is the application of summary operations. These operations compress each of the $m$ representations of $X_i$ ($\{T_{i,1}, T_{i,2}, \dots, T_{i,m}\}$) into a set of features through the application of statistical functions. A summary operation can be defined as follows:

\begin{align*}
  \text{Summary}_k: \mathbb{R}^q &\to \mathbb{R}\\
  T_{i,j} &\mapsto s_{i,j,k}
\end{align*}

\noindent where $\text{Summary}_k$ denotes the $k$-th summary operation, and $s_{i,j,k}$ denotes the feature obtained when applying the $k$-th summary operation to the $j$-th representation of the $i$-th embedding vector. 
Each $s_{i,j,k}$ is part of the set $S_{i,j}$, which represents the features describing $T_{i,j}$.

Essentially, each summary operation compresses a numeric vector into a scalar which summarises the current state of the time series in some way. A simple example is the arithmetic mean, which describes the central tendency.

\subsection{Feature Selection}\label{subsec:selection}

The feature generation process described above produces a large number of features. This procedure may lead to a problem of high dimensionality and, consequently, overfitting. We introduce a feature selection procedure to cope with this issue.

Depending on the nature of the time series, the features extracted may have three problems: (i) they may not be applicable, which leads to missing values; (ii) they may not vary enough across the observations and do not provide any information for forecasting; or (iii) they may be highly correlated with each other. Concerning the first problem, we remove any feature with more than a certain percentage of missing values, $na\_perc$. Features with a lower percentage of missing values are imputed using the median function. The second issue is dealt with by removing features with a low number of unique values. Specifically, we remove any feature whose number of unique values relative to the total number of observations is below $u\_perc$.
Finally, we apply a filter for removing correlated features. If a pair of features shows a level of correlation above $corr\_perc$, one of them is discarded.

This process leads to the final set of features, $Z_i$. We concatenate this set with the original embedding vector $X_i$. 

\section{Experiments}\label{sec:exp}

This section presents the empirical experiments carried out to validate the proposed approach. First, we detail the transform and summary operations used in the feature generation process of \texttt{VEST} (Section~\ref{sec:vectormrs_setup}). Then, we present the experimental setup (Section~\ref{sec:edesign}), describing the research question, case study, methods and respective hyper-parameters, and evaluation approach. We compare the proposed approach with state of the art approaches in Section~\ref{sec:results}. We perform a feature important analysis in Section \ref{sec:featimp}. Finally, we analyse the impact of sample size in the results obtained (Section \ref{sec:ssize}).

\subsection{\texttt{VEST} Setup}\label{sec:vectormrs_setup}

\subsubsection{Transform and Summary Operations}\label{sec:mapop}

The transform operations applied to each embedding vector $X_i$ are described in Table~\ref{tbl:transformops} and the summary operations applied to each representation $T_{i,j}$ are described in Table~\ref{tbl:summaryops}.

\begin{table}[!ht]
\centering
\caption{Transform operations used in \texttt{VEST}.}\label{tbl:transformops}
\begin{tabularx}{\textwidth}{cX}
\textbf{Operation} & \textbf{Description} \\ \hline
\textbf{I}   & The Identity transformation, in which each $X$ is mapped onto itself \\ \hline

\textbf{SMA}	& We apply a one-sided simple moving average which can be beneficial to smooth out spurious fluctuations and highlight the general trend. The number of periods is equal to the square root of the length of $X$, rounded to the nearest unit \\ \hline
\textbf{DIFF}	& First differences are applied to transform the original embedding vector into one without trend. This transformation can help with the modelling of time series with a strong trend component \\ \hline
\textbf{DIFF2}	& Second differences, which is equivalent to applying the \textbf{DIFF} operation twice to $X_i$. This transformation is useful for describing the curvature of the data  \\ \hline
\textbf{BC}	& Box-Cox transformation, for stabilising the variance of the time series. The transformation parameter is optimised using all the available observations according to Guerrero~\cite{guerrero1993time} (minimizing the coefficient of variation)  \\ \hline

\textbf{SIN}	& Sine terms of order 1 of the Fourier series. This transformation captures the seasonality of the time series. We remark that the frequency of the time series must be available to compute these terms \\ \hline

\textbf{COS}	& Similar and complementary to SIN, COS captures the cosine terms of order 1 of the Fourier series.  \\ \hline
\textbf{DWT} & We apply a 1-level discrete wavelet transform using the Daubechies wavelet~\cite{percival2006wavelet}, and retrieve the coefficients of the respective detail signal \\ \hline

\end{tabularx}
\end{table}

\begin{table}[!ht]
\centering
\caption{Summary operations used in \texttt{VEST}.}\label{tbl:summaryops}
\begin{tabularx}{\textwidth}{C{1.5cm}X}

\textbf{Operation} & \textbf{Description} \\ \hline
\textbf{MEAN} & Arithmetic mean, which is used to estimate the average level of the vector \\ \hline
\textbf{MDN} & Median: similar to the mean, but more robust to outliers \\ \hline
\textbf{SD} & Standard deviation, as a measure of the overall dispersion in the vector \\ \hline
\textbf{VAR} & Variance of the vector, which also measures dispersion \\ \hline
\textbf{IQR} & Inter-quartile range, which is another measure of dispersion of the data, but more robust to outliers  \\ \hline
\textbf{RD} & Relative Dispersion, which is estimated according to the ratio between the standard deviation of the vector and the standard deviation of the differenced vector~\cite{wang2006characteristic} \\ \hline
\textbf{MIN} & Minimum value of the vector \\ \hline
\textbf{MAX} & Maximum value of the vector \\ \hline
\textbf{LP} & Last known point of the vector \\ \hline
\textbf{SK} & Skewness of the distribution of the vector, which is a measure of its asymmetry~\cite{wang2006characteristic}  \\ \hline
\textbf{KRT} & Kurtosis for describing the flatness of the data with respect to a normal distribution~\cite{wang2006characteristic} \\ \hline
\textbf{P05}, \textbf{P95} & The 5th and 95th percentiles of the vector \\ \hline
\textbf{ACC\_1, ACC\_2} & Average (\textbf{ACC\_1}) and standard deviation (\textbf{ACC\_2}) of the acceleration of the vector, estimated according to the ratio between the simple moving average and the exponential moving average of equal period. In our experiments, the period for computing the moving averages was set to the squared root of the length of the vector, rounded to units \\ \hline
\textbf{BP} & Level of auto-correlation, which is estimated using a Box-Pierce test statistic~\cite{box1970distribution,wang2006characteristic} \\ \hline
\textbf{PACF} & Average value of the partial auto-correlation function of the vector up to 10 lags \\ \hline
\textbf{ACF} &  Average value of the auto-correlation function of the vector up to 10 lags \\ \hline
\textbf{LRD1} \textbf{LRD2} & Long-range dependence,
estimated using the Hurst exponent approach with wavelet transform with 1 (\textbf{LRD1}) and 2 moments (\textbf{LRD2})~\cite{wang2006characteristic} \\ \hline
\textbf{SLP} & Slope of the vector which describes its overall steepness~\cite{prudencio2004using} \\ \hline
\textbf{NORM} & Euclidean norm of the vector, which captures its total energy \\ \hline
\textbf{NO} & Number of outliers, estimated according to the number of observations above or below 1.5 times the inter-quartile range \\ \hline
\textbf{AMP} & Average amplitude of the fast Fourier transform of the vector \\ \hline
\textbf{STEP} & Binary random variable which denotes the presence of a step change~\cite{lemke2010meta}. This statistic detects structural breaks in the data \\ \hline
\textbf{PEAK\_I}, \textbf{PEAK\_D} & Number of local maxima (\textbf{PEAK\_I}) and local minima (\textbf{PEAK\_D}) in the vector~\cite{lemke2010meta}. These statistics describe the level of oscillation of the data \\ \hline
\textbf{OD} & Overall direction of the vector, estimated by the difference between the number of times the vector increases and the number of times the vector decreases \\ \hline
\textbf{PV\_ST, PV\_LT} & Short-term and long-term variability, respectively, estimated using the Poincaré plot~\cite{brennan2001existing} \\ \hline
\textbf{MLE} & Maximum Lyapunov exponent, which quantifies the chaotic level of a time series~\cite{wang2006characteristic} \\ \hline
\end{tabularx}
\end{table}

The set of transform operations used try to capture the dynamics of the time series from distinct perspectives. Moreover, the list of summary operations contains several statistics which try to capture different components; from centrality and dispersion to chaos and stochastic randomness. 

Overall, we apply eight different transformations and 32 different summary operations, leading to 256 different features before feature selection. Henceforth, we will employ the following notation to refer to a feature generated by \texttt{VEST}: \textit{TransformFunction.SummaryFunction}. For example \textbf{DIFF.MEAN} represents the average value of the embedding vector representation when transformed with the differencing operation.

\paragraph{Setup.} 

We set the $na\_perc$ value to 70. As such, we remove features which have more than 70\% of its values missing. Also, we set the $u\_perc$ threshold to 1. Therefore, we remove features where the percentage of unique values is below 1\% of the total number of observations. The feature correlation threshold ($corr\_perc$) was set to 95.

\subsection{Experimental Design}\label{sec:edesign}

The main research question addressed in this paper is the following:

\begin{center}
    \textit{Does \texttt{VEST}, an automatic feature engineering procedure, improve forecasting performance relative to a pure auto-regressive approach?}
\end{center}

\noindent Our experiments to answer this question can be split into the following items:
\begin{itemize}
    \item \textbf{RQ1}: Effect of \texttt{VEST} on the predictive performance of the state of the art pure auto-regressive approach. We assess the significance of results according to Bayesian methods;
    
    \item \textbf{RQ2}: Comparison of the forecasting performance relative to state of the art approaches, such as ARIMA and exponential smoothing;
    
    \item \textbf{RQ3}: Sensitivity to different learning algorithms;
    
    \item \textbf{RQ4}: Analysis of the different feature selection approaches;

    \item \textbf{RQ5}: Analysis of the importance scores of each transformation and summary operations (Section \ref{sec:vectormrs_setup});
    
    \item \textbf{RQ6}: Sensitivity to different time series sample size.
    
\end{itemize}

\subsubsection{Data}

We used time series from two sources. From the \textit{tsdl} benchmark library~\cite{tsdlpackage}, we selected all the univariate time series with at least 1000 observations and which have no missing values. This represents 55 time series. These show a varying sampling frequency (e.g. daily) and are from different domains of application. For a complete description of these time series, we refer to the their source~\cite{tsdlpackage}. We also included 35 time series used by Cerqueira et al.~\cite{cerqueira2019arbitrage}. Essentially, from the set of 62 series used by the authors, we selected those with at least 1000 observations and that were not originally from the \textit{tsdl} database (since these were already retrieved as described above). We refer to the work by Cerqueira et al.~\cite{cerqueira2019arbitrage} for a description of those time series.

\subsubsection{Parameter Setting}\label{sec:parset}

For each time series, we optimise the embedding dimension using validation data, testing values from 10 to 30. 
The chosen embedding dimension $p$ is the one minimising the error (Section \ref{sec:eval} describes the evaluation metric). In this analysis, we train a model according to pure auto-regressive forecasting models (i.e., no feature engineering is involved at this point). We set the minimum value for searching the embedding dimension to 10 to guarantee a reasonable number of observations for computing the transform and summary operations of \texttt{VEST}.

We focus on two learning algorithms. One is the cubist method \cite{Cubist2014}, which is a variant of the model tree proposed by Quinlan~\cite{quinlan1993combining};
This method is competitive in time-dependent data~\cite{ikonomovska2011learning,cerqueira2019arbitrage}. 
We also use the lasso~\cite{tibshirani1996regression} regression algorithm.
Each one of the methods was optimised according to a grid search using validation data.

We will present results that quantify the importance of each feature across the 90 problems. We resort to the RReliefF \cite{robnik1997adaptation} method for this task. RReliefF (for Regressional ReliefF) extends ReliefF for numerical prediction problems. It estimates the importance of each feature in a data set by measuring the variability of the values of the features in the neighbourhood the observations. 
This method has been shown to have a connection to impurity measures~\cite{robnik1997adaptation}.

\subsubsection{Methods}

The learning algorithms indicated above were trained according to the following procedures:

\begin{itemize}
    \item \texttt{AR}: A pure auto-regressive process, where the value of the next set of observations is predicted according to the most recent $p$ values. This is the typical approach to tackle time series forecasting problems;
    
    \item \texttt{AR+VEST}: The proposed approach -- the combination of \texttt{AR} with the features obtained with \texttt{VEST};
    
    \item \texttt{VEST}: A baseline which discards the \texttt{AR} component and models the future behaviour of the time series using only the features obtained with \texttt{VEST};
    
    \item \texttt{AR+BT}: Variant of \texttt{AR+VEST}, in which the feature selection approach is different: We use the feature from only a single representation. For each time series, we pick the transformation which maximizes feature importance (according to RReliefF). The importance scores are average across the available summary operations. In other words, this variant contains all the summary operations detailed on Table \ref{tbl:summaryops}, but are computed only for the best estimated transformation;
    
    \item \texttt{AR+BF}: Another variant of \texttt{AR+VEST}, in which we select a single transformation for summary operation. This is similar to the variant \texttt{AR+BT} described above. The difference is that, in this case, the single transformation is picked for each summary operation. This selection is also based on feature importance. To be precise, for each time series and for each summary operation, we select the transformation that maximizes feature importance.
    
\end{itemize}

Additionally, we also include \texttt{ARIMA}, \texttt{ETS}, and \texttt{TBATS} in the experimental setup. These methods are state-of-the-art approaches for time series forecasting. They establish a reference to assess whether the results obtained here are acceptable or not. We resort to the implementations provided by the forecast R package~\cite{forecast}, which automatically tunes these methods to an optimal parameter setting.

\subsubsection{Evaluation}\label{sec:eval}

We use a holdout repeated in multiple testing periods as the estimation method according to \cite{cerqueira2019performanceestimation}. We perform 10 repetitions of this procedure. The training size in each repetition was set to 60\% of the total number of observations, while the subsequent 10\% of observations are used for testing. In each repetition, part of the training data (also 10\% of it) was used as a validation set to optimise parameters, such as the embedding dimension or the parameters of the learning algorithms.

Regarding the evaluation metric, we use the mean absolute scaled error (MASE), which is a typical measure of forecasting performance \cite{hyndman2006another}. We average the loss of each method across the repetitions of the holdout procedure described above. 
We evaluate the statistical significance of the results according to a Bayesian analysis~\cite{benavoli2017time}. In particular, we applied the Bayes sign test to compare pairs of methods across multiple problems. In the next section, we specify the setup of the test. For a thorough read on Bayesian analysis for comparing predictive models, we refer to the work by Benavoli et al.~\cite{benavoli2017time}. 

\subsection{Results}\label{sec:results}

In order to have a commensurable metric across data sets, we compute the percentage difference between the MASE of each approach and a benchmark model. We use \texttt{AR+VEST} as the benchmark as it represents the proposed approach that combines auto-regression with automatic feature engineering.
We formalise the percentage of difference computation as follows:

\begin{equation}
    \frac{L_a - L_{\texttt{AR+VEST}}}{L_{\texttt{AR+VEST}}} * 100
\end{equation}

\noindent where $L_{\texttt{AR+VEST}}$ and $L_{a}$ represent the loss of the model \texttt{AR+VEST} and the loss of model $a$ (the one is under comparison), respectively. We perform a Bayesian analysis of the results using the Bayes sign test~\cite{benavoli2017time}. We define the \textit{region of practical equivalence}~\cite{benavoli2017time} (ROPE) to be the interval [-2.5, 2.5]. Essentially, this means that the performance level of these two methods are nearly indistinguishable if the percentage difference in predictive performance between them falls within this interval.

\begin{figure}[ht]
\centering
\includegraphics[width=.9\textwidth]{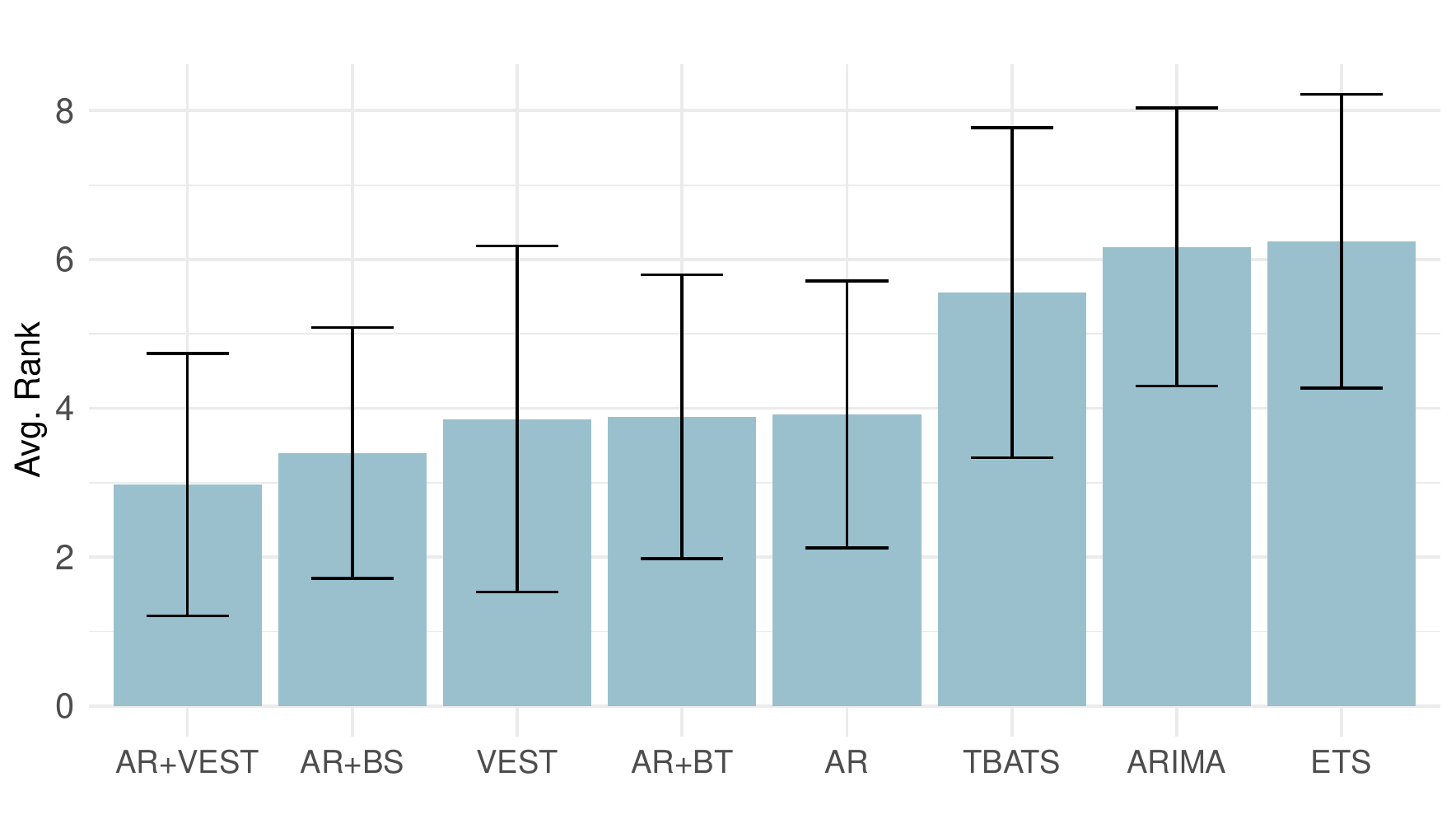}
\caption{The average rank, and respective standard deviation, of each method across the 90 time series when using cubist as learning algorithm.}
\label{fig:avgrankall}
\end{figure}

We start by analysing the average rank, and respective standard deviation, of each method. This is reported in Figure \ref{fig:avgrankall} using cubist as learning algorithm. A method with rank 1 in a task means that it was the best performing one in that task. The average rank describes the average position of each method relative to the remaining ones. \texttt{AR+VEST} shows the best average rank score, which shows the usefulness of the proposed approach.

\begin{figure}[ht]
\centering
\includegraphics[width=\textwidth]{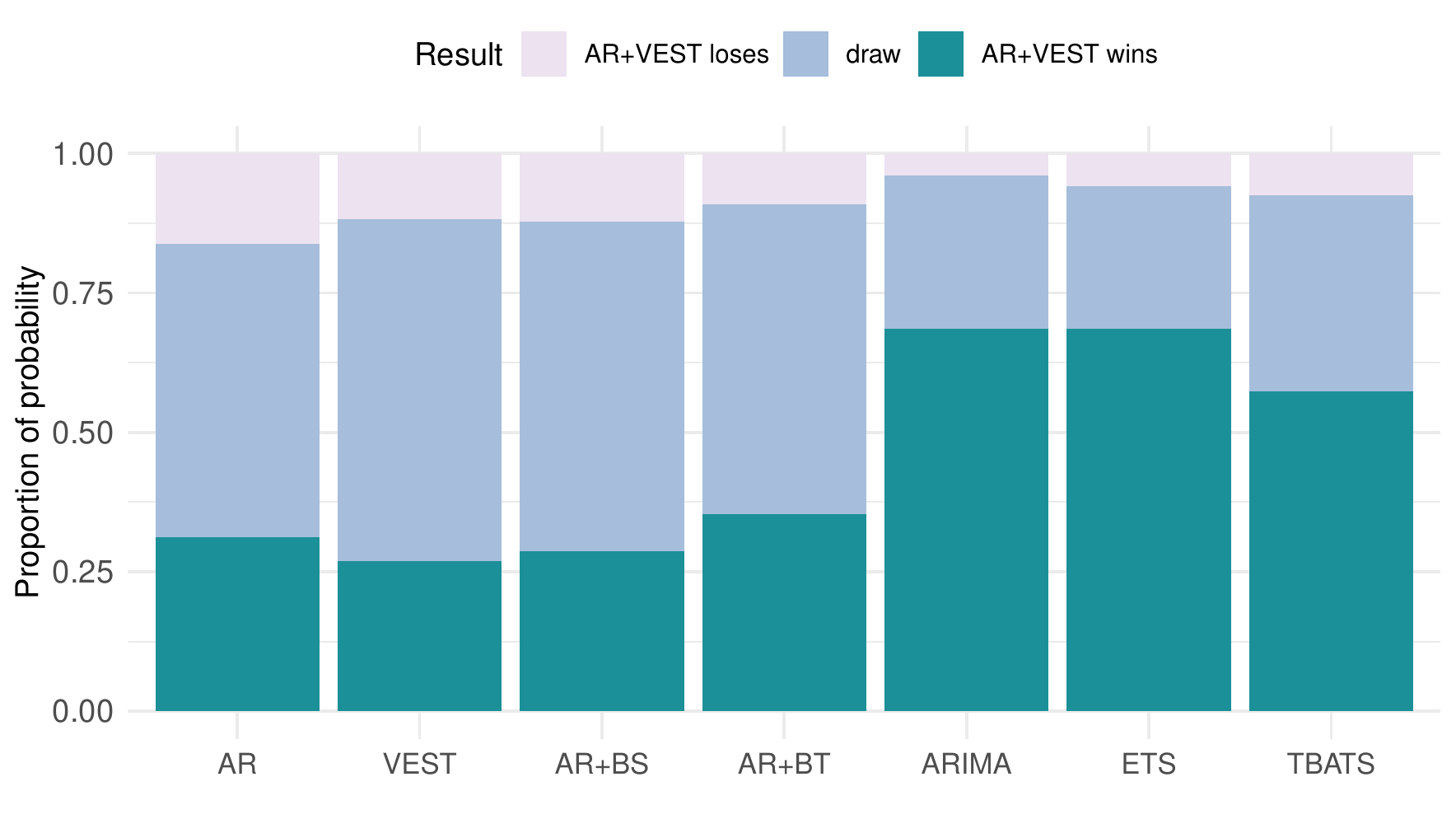}
\caption{Probability of each method winning, drawing, or losing significantly against \texttt{AR+VEST} (the proposed method) according to the Bayes sign test. Results shown for the cubist method.}
\label{fig:bs_all}
\end{figure}

In terms of significance analysis, Figure \ref{fig:bs_all} shows the probability that the respective method wins, draws (result within the ROPE), or loses significantly, against the proposed model (\texttt{AR+VEST}) also when using the cubist learning algorithm. \texttt{AR+VEST} significantly outperforms the standard auto-regressive model (\texttt{AR}) with around 30\% probability (\textbf{RQ1}). The opposite scenario occurs with around 17\% probability. In the remaining cases, the results are within the ROPE, which means the approaches are statistically equivalent. \texttt{AR+VEST} is also significantly better relative to state of the art forecasting approaches, including \texttt{ARIMA}, \texttt{TBATS}, and \texttt{ETS} (\textbf{RQ2}). This corroborates previous experiments shown by Cerqueira et al. \cite{cerqueira2019arbitrage}.

These results show important evidence that feature-based forecasting is worthwhile, and may be important to improve forecasting performance.

\begin{figure}[ht]
\centering
\includegraphics[width=.9\textwidth]{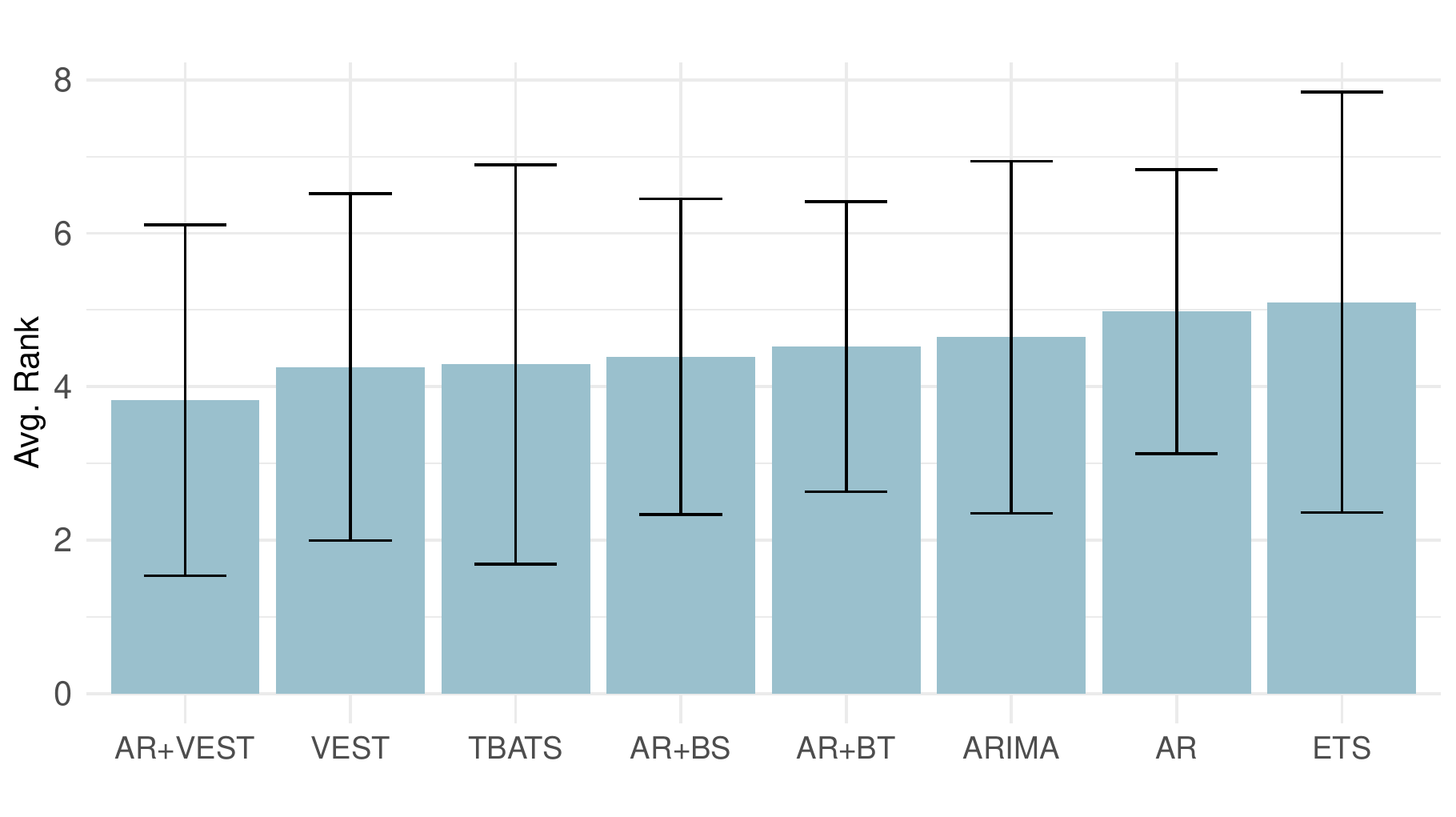}
\caption{The average rank, and respective standard deviation, of each method across the 90 time series.}
\label{fig:avgrankall_lasso}
\end{figure}

\begin{figure}[ht]
\centering
\includegraphics[width=.9\textwidth]{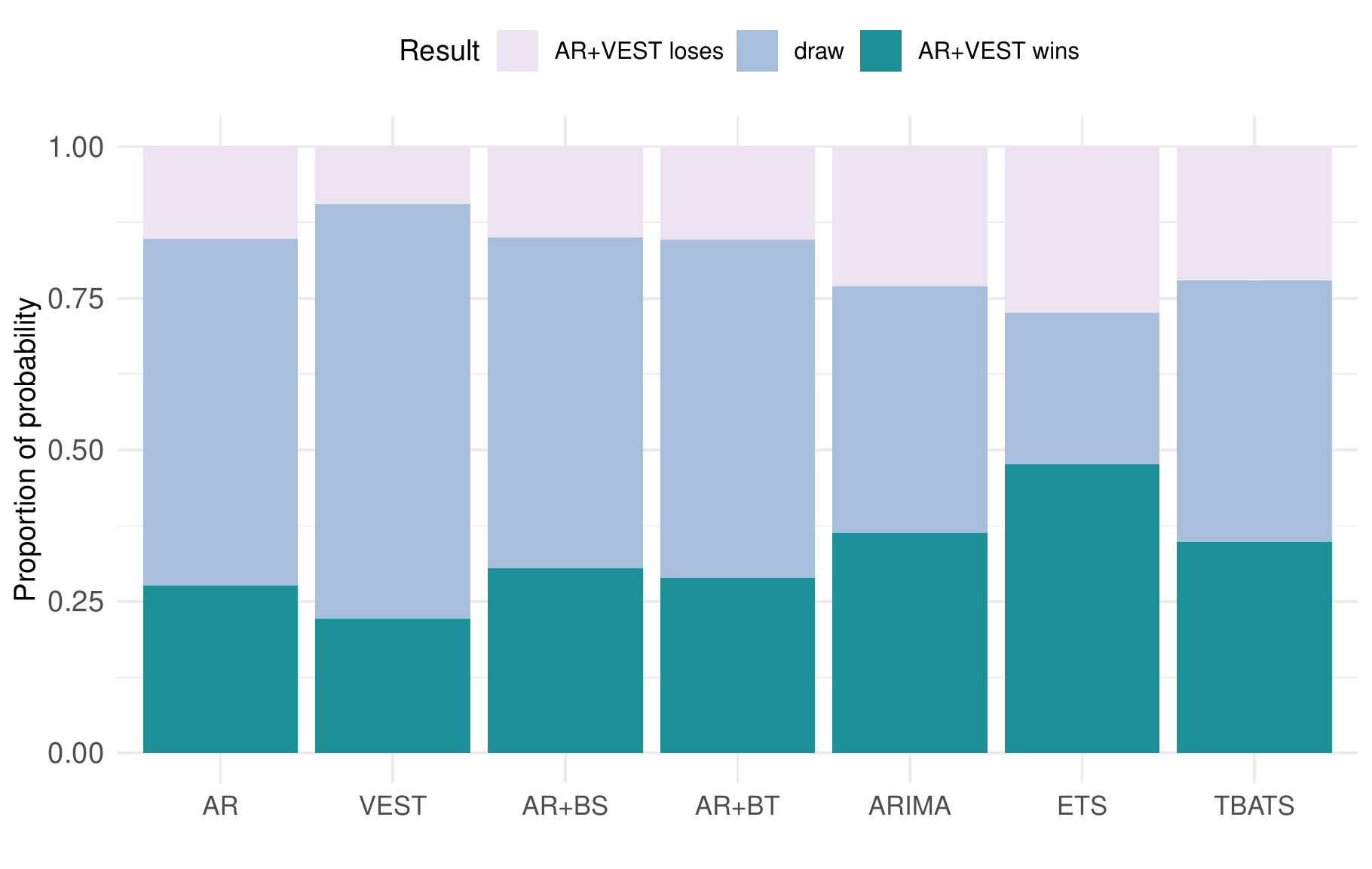}
\caption{Probability of each method winning, drawing, or losing significantly against \texttt{AR+VEST} (the proposed method) according to the Bayes sign test. Results show for the lasso method.}
\label{fig:bs_all_lasso}
\end{figure}

Figures \ref{fig:avgrankall_lasso} and \ref{fig:bs_all_lasso} are similar to Figures Figures \ref{fig:avgrankall} and \ref{fig:bs_all}, but the analysis is carried out using the lasso learning algorithm. Although not identical, the illustration shows that performance gains are also obtained when using this method (\textbf{RQ3}).

Besides state of the art forecasting approaches, the results also indicate that \texttt{AR+VEST} outperforms three variants: \texttt{VEST}, \texttt{AR+BT} and \texttt{AR+BS}. \texttt{VEST} denotes the approach that discards the auto-regressive attributes and uses only the features derived from the proposed framework to forecast the next value of the time series. However, the results show that combining \texttt{AR} with \texttt{VEST} is critical to the performance obtained. By itself, \texttt{VEST} shows a competitive performance, but does not provide a consistent advantage. 

Regarding \texttt{AR+BT} and \texttt{AR+BS}, these variants provide a different approach for selecting the features from \texttt{VEST}. We devised \texttt{AR+BT} (the approach which selects a single transformation for each time series) to show the usefulness of multiple representations in a given problem. On the other hand, by outperforming \texttt{AR+BS} (the approach that uses a single transformation per summary operation), we show that multiple transformations are useful for a given summary operation even within the same problem. We dealt with eventual redundancies with a simple correlation filter, as described in the experimental design (\textbf{RQ4}). 

\subsection{Feature Importance}\label{sec:featimp}

In the previous section, we presented significant empirical evidence that the use of \texttt{VEST} can significantly improve forecasting performance.
In this section, we dive deeper into this matter by analysing the importance of the features used in the development of models. This covers research question \textbf{RQ5}.

\subsubsection{Rank Distributions}

We start by analysing the distribution of the rank of the feature importance across the 90 time series. We proceed as follows. 

\begin{enumerate}
    \item We measure the RReliefF of each feature. This score is averaged across the repetitions of the repeated holdout procedure;
    \item We compute the rank of each feature according to its score of importance across the 90 time series. A feature with rank 1 in a given time series has the best score of importance in that problem. We split the computation of ranks into three parts according to the following criteria:
    \begin{itemize}
        \item All operations: We compute the rank of all features irrespective of the underlying representation;
        \item Representation:
        We compute the rank of each transformation. Specifically, we average the score of the importance of the features for each representation. For example, the average importance of all features using the DIFF representation. In this analysis, we include the importance of the past lags of the series (denoted as LAG variables);
        \item Summary function: We also compute the rank of each summary operation. Similarly as above, we average the score of importance across summary operation to obtain the overall importance of the respective function.
\end{itemize}

\end{enumerate}

\paragraph{Overall Rank.}

\begin{figure}[t]
\centering
\includegraphics[width=\textwidth]{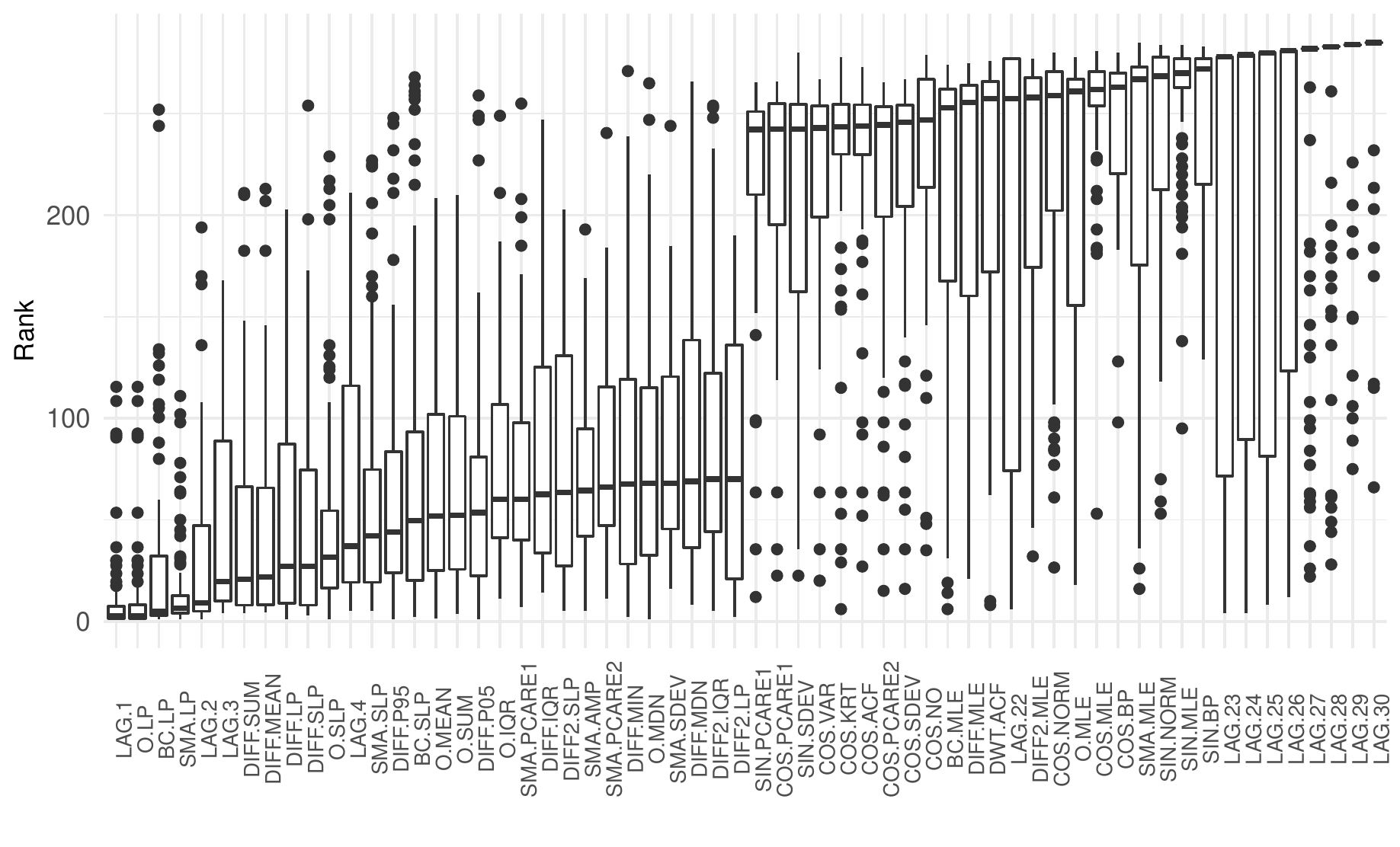}
\caption{Distribution of rank importance of the top 30 and bottom 30 features. The rank of a particular feature in a given problem is computed according to its importance.}
\label{fig:fi_rankdist}
\end{figure}

Figure~\ref{fig:fi_rankdist} shows the results of the overall rank as a set of boxplots (one for each feature), which are ordered by median importance rank (lower values are better). The names of the features (x-axis) follows the convention described before. 
In the interest of conciseness, we only show the top and bottom 30 features in terms of median rank. 

The feature with the best median rank is \textbf{LAG.1}, which represents the last known value of the time series in a given point in time. Figure~\ref{fig:fi_rankdist} clearly shows the advantage of methods to systematically generate large numbers of new features, when compared to, typically, a few features generated manually, based on domain knowledge. We observe that, overall, there is a great dispersion in the rank importance, showing that different features are more important in different time series. In fact, even those features with high median rank are among the most important in some of the problems.

\begin{figure}[t]
\centering
\includegraphics[width=.8\textwidth]{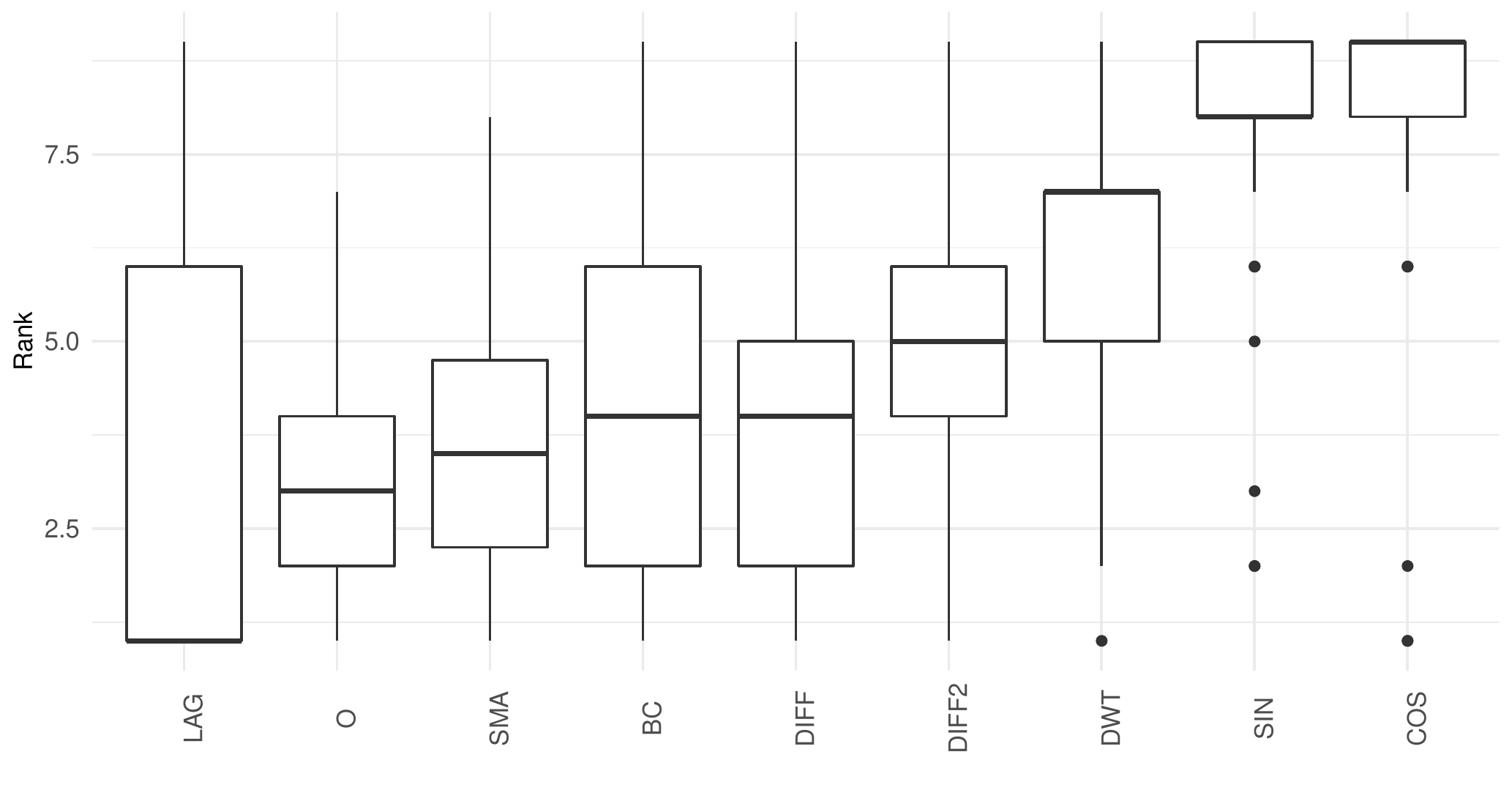}
\caption{Distribution of rank importance for each type of feature.}
\label{fig:fi_rankdist_map}
\end{figure}

\paragraph{Rank by Representation.} Figure \ref{fig:fi_rankdist_map} provides a similar analysis as Figure \ref{fig:fi_rankdist}, but combines the results by representation, as explained above. 
The boxplots provide additional evidence for two observations made previously. Firstly, it shows that, although the obtained features from \texttt{VEST} improve predictive performance, the previous points of the time series (LAG features) provide useful information. In fact, they obtain the best median rank. Secondly, the high dispersion of the rank distributions shows that there is no particular representation which is the most appropriate for all time series. This provides additional evidence of the usefulness and complementarity of the different representations, as observed earlier.

\paragraph{Rank by Summary Operation.}

\begin{figure}[ht]
\centering
\includegraphics[width=\textwidth]{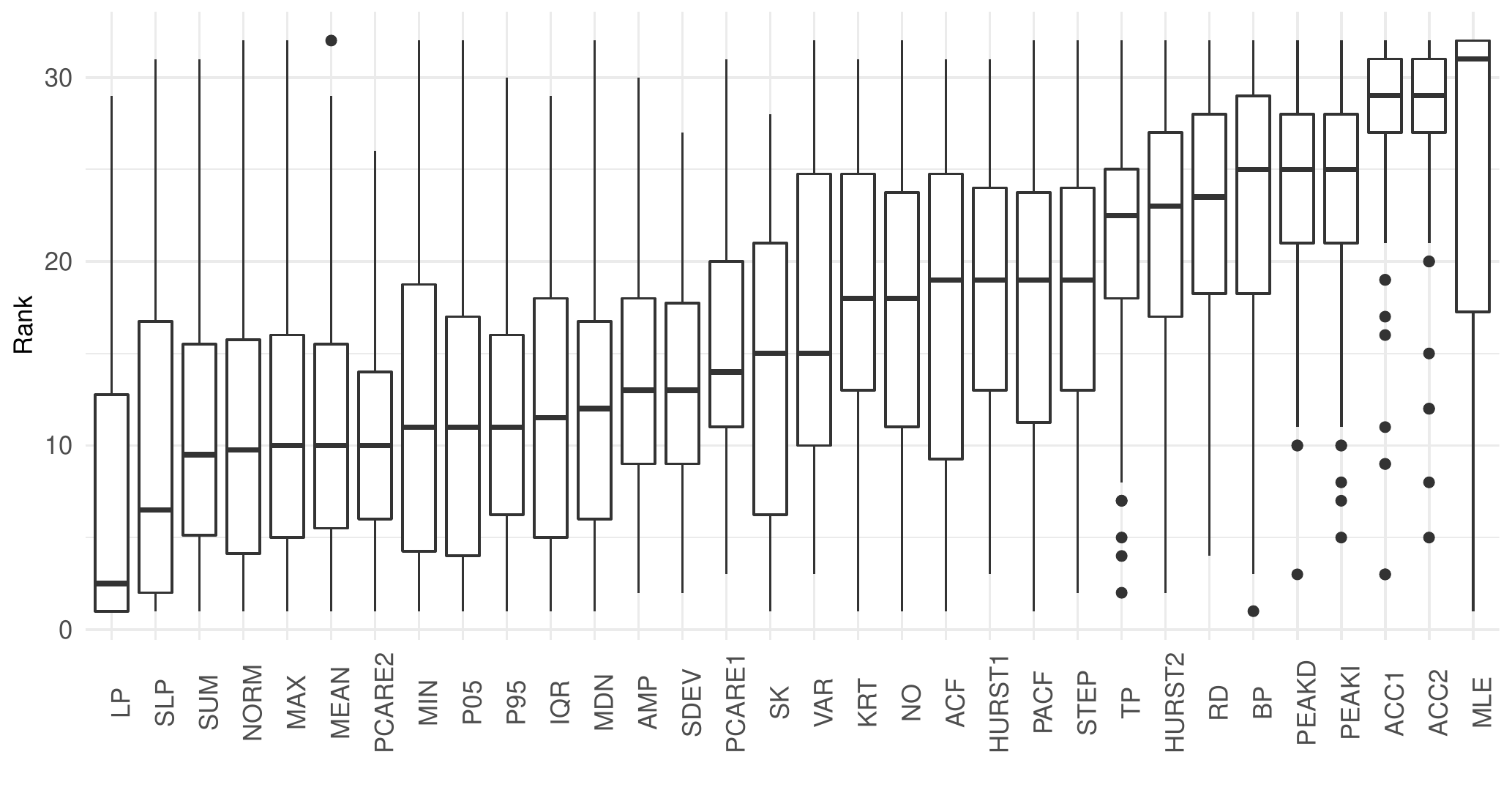}
\caption{Distribution of rank importance for each summary operation.}
\label{fig:fi_rankdist_stat}
\end{figure}

Figure \ref{fig:fi_rankdist_stat} shows a similar analysis but referring to each summary operation. Again, the boxplots show high dispersion suggesting that different statistics are more valuable in different tasks. The statistic with the best median rank is LP, which denotes the last point of the respective transformation.

\subsection{Impact of Sample Size}\label{sec:ssize}

We focused the experimental setup on high frequency time series. This type of data sets is increasingly relevant in many practical applications due to the widespread adoption of sensors. High frequency time series are typically associated with larger sample sizes relative to lower frequency ones. 
We hypothesise that the sample size is important when performing feature engineering with a method such as \texttt{VEST}. In a small data set additional attribute variables may lead to over-fitting issues due to the \textit{curse of dimensionality}. Therefore, it is important to collect a reasonable amount of observations for feature engineering. 

We test the hypothesis above by repeating the experiments with increasing sample size values, similarly to Cerqueira et al. \cite{cerqueira2019machine} (\textbf{RQ6}). To be more precise, we start by truncating the sample size of the time series to 3000 observations. Only 42 of the 90 time series had enough sample size, and we focus this analysis on that subset of problems. Afterwards, we repeated the experiments (as described in Section \ref{sec:edesign}) in 30 different samples sizes from 100 to 3000 in steps of 100 observations ($\{100, 200, \dots, 3000\}$). We remark that, in this particular experiment, we focus on a simple holdout estimation method in which 80\% of the initial observations are used for training and the subsequent 20\% data points are used for testing. 
Accordingly, we evaluate the performance of \texttt{AR+VEST} relative to other approaches which do not use \texttt{VEST}. In this experiment, we remove the variants of the proposed method, and keep only \textbf{AR}, \textbf{ETS}, \textbf{TBATS}, and \textbf{ARIMA}. The performance is evaluated as follows: for each problem as for each sample size we compute the MASE error of each approach. Then, each method is ranked according to this error (lower error gives lower rank). We average the rank of each method across the 42 time series in each sample size experiment. This allows us to describe how the average rank of each method evolves as the sample size increases. We remark that we resort to the rank, as opposed to the MASE loss, because it is non-parametric and robust to outliers. Finally, we remark that we focus on the cubist learning algorithm for this analysis in the interest of conciseness. The results are similar when using the lasso algorithm.

\begin{figure}[ht]
\centering
\includegraphics[width=\textwidth]{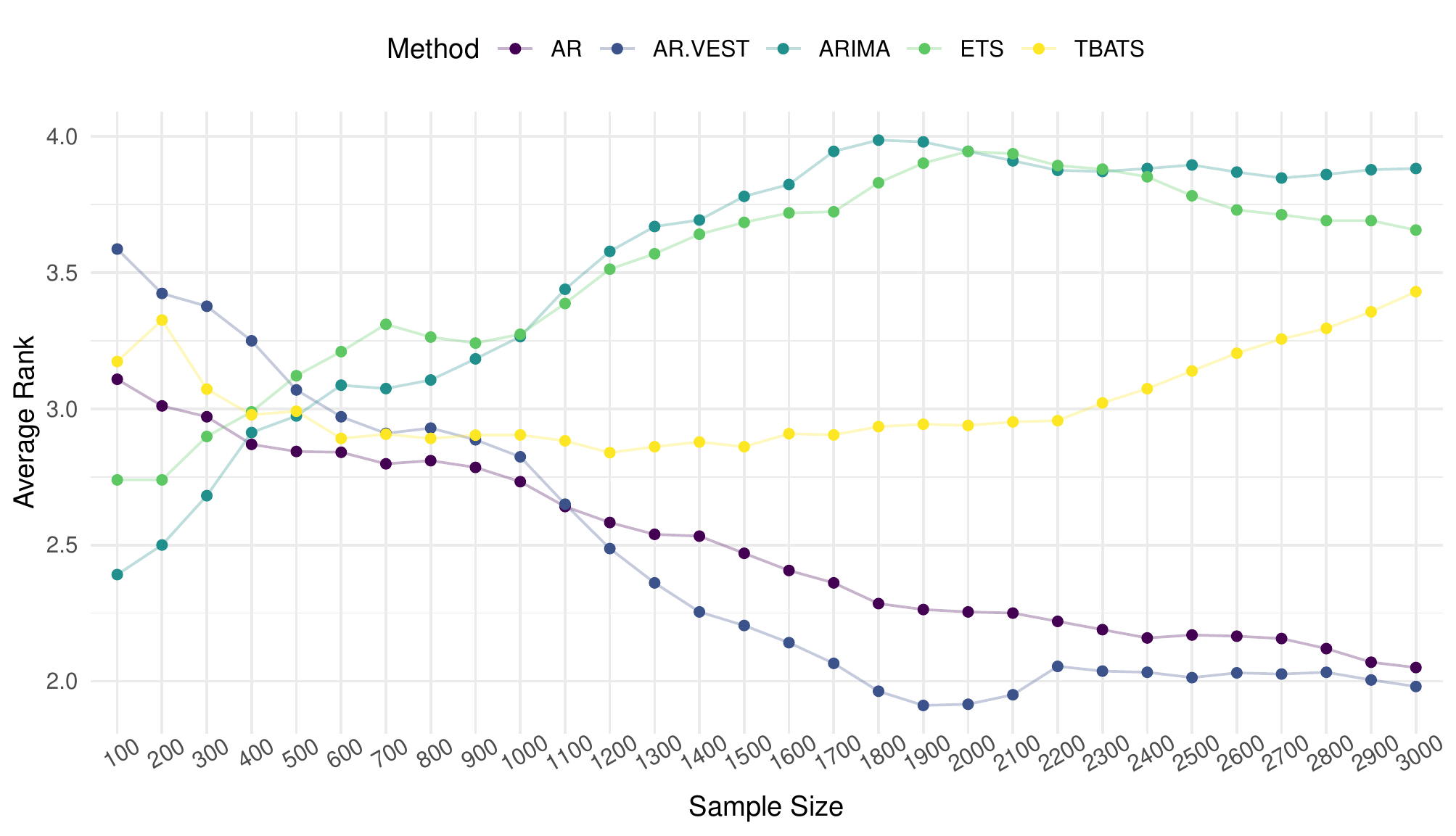}
\caption{The average rank (computed across the 42 time series) of each method as the sample size of the time series increases. Lower values are better}
\label{fig:lc}
\end{figure}

The results are presented in Figure \ref{fig:lc}, which shows the average rank of each method across the 42 time series with an increasing sample size. The results show that, when the sample size is small, \texttt{AR+VEST} shows worse results than all other methods, including \texttt{AR} and the state of the art forecasting methods \texttt{ARIMA}, \texttt{ETS}, and \texttt{TBATS}. However, as the sample size increases, \texttt{AR+VEST} becomes the approach with best average rank. We remark that the average rank scores may be slightly different from the analysis shown previously as there are only 42 time series under analysis in this scenario.

\section{Discussion}\label{sec:disc}

The experiments carried out in the previous section show the benefits of using \texttt{VEST} for time series forecasting tasks. In this section, we discuss the results obtained and point future directions of research.

\subsection{Main Results}

\texttt{VEST} is a framework for automatically extracting relevant features from the embedding vectors representing the time series.
We showed the usefulness of \texttt{VEST} to tackle time series forecasting tasks based on an extensive set of experiments. 
When the features generated by \texttt{VEST} are combined with a state of the art auto-regressive model (\texttt{AR}), forecasting performance significantly improves relative to only using \texttt{AR}.

We explored these results from different perspectives. Particularly, we presented an analysis which suggests that there is no specific representation or summary statistics which is more appropriate for all time series problems. Even within a single time series, the results suggest that applying summary operations to different representations is important for forecasting performance.
This outcome shows the potential benefit of using an automatic approach to extract meaningful features for this type of data. 
Rather than finding a single feature that improves results across multiple problems, \texttt{VEST} obtains a set of features, each one of which is very important for a small, particular subset of the problems, although not very relevant on the remaining ones.


We believe our work is relevant for automated machine learning frameworks, especially to enable professional with low technical skill to develop accurate forecasting models efficiently \cite{taylor2018forecasting}. 

\subsection{Points for Improvement}

Despite significant gains in forecasting performance, we believe it is possible to improve the proposed feature engineering process. 

\texttt{VEST} is designed as a brute force approach. It works by testing different representations, which are then summarised using different statistics. Those with low feature importance are removed using a feature selection filter. The key factor for the gains in performance is the predictive quality of the features that are tested. 
In this context, a potentially interesting research line is to develop a method for selecting apriori which transform and summary operations should be computed. For example, Reis \cite{do2019automated} attempts to use meta-learning to predict whether a given feature is going to improve predictive performance. A similar approach could be developed for extending \texttt{VEST}. Other possible interesting solutions are landmarkers \cite{Pfahringer2000} or bayesian optimisation \cite{rasmussen2003gaussian}. Notwithstanding, we remark that, in the proposed framework, the different transform operations are independent of each other, and so are summary ones. Therefore, the processes within each step can run in parallel. 

Another point of improvement for \texttt{VEST} is long-term analysis. Except for the dynamic harmonic regression procedure used to estimate the Fourier terms, \texttt{VEST} focuses on extracting information from the past $p$ lags. In other words, feature extraction is self-contained within each embedding vector. In future work, we plan to extend the approach to include a longer-term analysis and extract information across embedding vectors. Such analysis enables a more long-term perspective on the dynamics of the time series. An example of a long-term feature is one attempting to capture a ``number of observations since an outlier occurred".

Although the pool of operations applied cover many properties of time series, the set of transform and summary operations can be increased. 
Other transformations could be carried out, for example, seasonal adjustment or discrete cosine transform. One could also combine available transform operations, e.g. a transformation which applies the operations BC (Box-Cox transformation) and DIFF (first differences transformation) sequentially (c.f. Section \ref{sec:mapop}). 

As we mentioned previously, \texttt{VEST} is designed to extract endogenous features from time series. Notwithstanding, external information may be crucial for building accurate predictive models. With additional time series as explanatory variables, the number of operations to be carried out may be too high for a brute force approach. Thus, the ideas outlined in the second paragraph of this section may be important in these scenarios.

Our experiments are based on 90 time series with a high sampling frequency (daily or higher). Research is still necessary to show the impact of feature engineering in time series with lower sampling frequency. In Section \ref{sec:ssize}, we showed that time series sample size is important for the proposed feature engineering solution. 

Another point for improvement for \texttt{VEST} is multi-step forecasting. During our experiments, we did not find enough evidence that \texttt{VEST} may be better than \texttt{AR} for multi-step forecasting. We intend to explore this topic in future research.

\section{Summary}\label{sec:conclusions}

Time series forecasting is a relevant predictive task in many domains of application. Data-driven organisations rely on forecasting systems to cope with future uncertainty and support their decision-making process.

One of the most important tasks in machine learning is feature engineering. However, this task still requires considerable manual effort and expertise from practitioners \cite{kaul2017autolearn}. This lead to increasing demand for approaches that automate this part of machine learning projects.

In this work, we present a novel automatic procedure for feature engineering using time series data. The proposed approach, called \texttt{VEST}, is specifically designed to tackle time series forecasting problems. \texttt{VEST} is based on the manipulation of the embedding vectors, which represent the past recent observations used to predict the future ones. It works by transforming time series sub-sequences into distinct representations. Describing a time series using multiple representations may be useful for capturing its dynamics from different perspectives. Each representation is summarised using statistical functions, such as the mean. After a feature selection process, the final set of features across the available representations is coupled with an auto-regressive model.

We validated the proposed approach using an extensive set of experiments, comprised of 90 time series from several domains of application. The results show that the features provided by \texttt{VEST}, along with auto-regression, lead to significant gains in forecasting performance. 

In future work, we will extend the approach to other forecasting scenarios, for example, multivariate time series or multi-step forecasting. \texttt{VEST} is publicly available as an \textit{R} software package.

\begin{acknowledgements}

This work is financed by National Funds through the Portuguese funding agency, FCT - Fundação para a Ciência e a Tecnologia, within project UIDB/50014/2020.

\end{acknowledgements}

\bibliographystyle{spmpsci}      

\begin{thebibliography}{10}
\providecommand{\url}[1]{{#1}}
\providecommand{\urlprefix}{URL }
\expandafter\ifx\csname urlstyle\endcsname\relax
  \providecommand{\doi}[1]{DOI~\discretionary{}{}{}#1}\else
  \providecommand{\doi}{DOI~\discretionary{}{}{}\begingroup
  \urlstyle{rm}\Url}\fi

\bibitem{barandas2020tsfel}
Barandas, M., Folgado, D., Fernandes, L., Santos, S., Abreu, M., Bota, P., Liu,
  H., Schultz, T., Gamboa, H.: Tsfel: Time series feature extraction library.
\newblock SoftwareX \textbf{11}, 100,456 (2020)

\bibitem{benavoli2017time}
Benavoli, A., Corani, G., Dem{\v{s}}ar, J., Zaffalon, M.: Time for a change: a
  tutorial for comparing multiple classifiers through bayesian analysis.
\newblock The Journal of Machine Learning Research \textbf{18}(1), 2653--2688
  (2017)

\bibitem{box2015time}
Box, G.E., Jenkins, G.M., Reinsel, G.C., Ljung, G.M.: Time series analysis:
  forecasting and control.
\newblock John Wiley \& Sons (2015)

\bibitem{box1970distribution}
Box, G.E., Pierce, D.A.: Distribution of residual autocorrelations in
  autoregressive-integrated moving average time series models.
\newblock Journal of the American statistical Association \textbf{65}(332),
  1509--1526 (1970)

\bibitem{brennan2001existing}
Brennan, M., Palaniswami, M., Kamen, P.: Do existing measures of poincare plot
  geometry reflect nonlinear features of heart rate variability?
\newblock IEEE transactions on biomedical engineering \textbf{48}(11),
  1342--1347 (2001)

\bibitem{cerqueira2019performanceestimation}
Cerqueira, V., Torgo, L., Mozetic, I.: Evaluating time series forecasting
  models: an empirical study on performance estimation methods.
\newblock Mach Learn (2020)
\newblock https://doi.org/10.1007/s10994-020-05910-7

\bibitem{cerqueira2017dets}
Cerqueira, V., Torgo, L., Oliveira, M., Pfahringer, B.: Dynamic and
  heterogeneous ensembles for time series forecasting.
\newblock In: 2017 IEEE International Conference on Data Science and Advanced
  Analytics (DSAA), pp. 242--251 (2017).
\newblock \doi{10.1109/DSAA.2017.26}

\bibitem{cerqueira2019arbitrage}
Cerqueira, V., Torgo, L., Pinto, F., Soares, C.: Arbitrage of forecasting
  experts.
\newblock Machine Learning \textbf{108}(6), 913--944 (2019)

\bibitem{cerqueira2019machine}
Cerqueira, V., Torgo, L., Soares, C.: Machine learning vs statistical methods
  for time series forecasting: Size matters.
\newblock arXiv preprint arXiv:1909.13316  (2019)

\bibitem{chatfield2000time}
Chatfield, C.: Time-series forecasting.
\newblock CRC Press (2000)

\bibitem{christ2016distributed}
Christ, M., Kempa-Liehr, A.W., Feindt, M.: Distributed and parallel time series
  feature extraction for industrial big data applications.
\newblock arXiv preprint arXiv:1610.07717  (2016)

\bibitem{esling2012time}
Esling, P., Agon, C.: Time-series data mining.
\newblock ACM Computing Surveys (CSUR) \textbf{45}(1), 12 (2012)

\bibitem{fulcher2017hctsa}
Fulcher, B.D., Jones, N.S.: hctsa: A computational framework for automated
  time-series phenotyping using massive feature extraction.
\newblock Cell systems \textbf{5}(5), 527--531 (2017)

\bibitem{guerrero1993time}
Guerrero, V.M.: Time-series analysis supported by power transformations.
\newblock Journal of Forecasting \textbf{12}(1), 37--48 (1993)

\bibitem{guyon2006introduction}
Guyon, I., Elisseeff, A.: An introduction to feature extraction.
\newblock In: Feature extraction, pp. 1--25. Springer (2006)

\bibitem{tsdlpackage}
Hyndman, R., Yang, Y.: tsdl: Time Series Data Library (2019).
\newblock Https://finyang.github.io/tsdl/, https://github.com/FinYang/tsdl

\bibitem{forecast}
Hyndman, R.J., with contributions~from George~Athanasopoulos, Razbash, S.,
  Schmidt, D., Zhou, Z., Khan, Y., Bergmeir, C., Wang, E.: forecast:
  Forecasting functions for time series and linear models (2014).
\newblock {R} package version 5.6

\bibitem{hyndman2006another}
Hyndman, R.J., et~al.: Another look at forecast-accuracy metrics for
  intermittent demand.
\newblock Foresight: The International Journal of Applied Forecasting
  \textbf{4}(4), 43--46 (2006)

\bibitem{ikonomovska2011learning}
Ikonomovska, E., Gama, J., D{\v{z}}eroski, S.: Learning model trees from
  evolving data streams.
\newblock Data mining and knowledge discovery \textbf{23}(1), 128--168 (2011)

\bibitem{kahn2003measure}
Kahn, K.B.: How to measure the impact of a forecast error on an enterprise?
\newblock The Journal of Business Forecasting \textbf{22}(1), 21 (2003)

\bibitem{kang2017visualising}
Kang, Y., Hyndman, R.J., Smith-Miles, K.: Visualising forecasting algorithm
  performance using time series instance spaces.
\newblock International Journal of Forecasting \textbf{33}(2), 345--358 (2017)

\bibitem{kanter2015deep}
Kanter, J.M., Veeramachaneni, K.: Deep feature synthesis: Towards automating
  data science endeavors.
\newblock In: 2015 IEEE International Conference on Data Science and Advanced
  Analytics (DSAA), pp. 1--10. IEEE (2015)

\bibitem{katz2016explorekit}
Katz, G., Shin, E.C.R., Song, D.: Explorekit: Automatic feature generation and
  selection.
\newblock In: 2016 IEEE 16th International Conference on Data Mining (ICDM),
  pp. 979--984. IEEE (2016)

\bibitem{kaul2017autolearn}
Kaul, A., Maheshwary, S., Pudi, V.: Autolearn—automated feature generation
  and selection.
\newblock In: 2017 IEEE International Conference on Data Mining (ICDM), pp.
  217--226. IEEE (2017)

\bibitem{keogh2004towards}
Keogh, E., Lonardi, S., Ratanamahatana, C.A.: Towards parameter-free data
  mining.
\newblock In: Proceedings of the tenth ACM SIGKDD international conference on
  Knowledge discovery and data mining, pp. 206--215 (2004)

\bibitem{khurana2016cognito}
Khurana, U., Turaga, D., Samulowitz, H., Parthasrathy, S.: Cognito: Automated
  feature engineering for supervised learning.
\newblock In: 2016 IEEE 16th International Conference on Data Mining Workshops
  (ICDMW), pp. 1304--1307. IEEE (2016)

\bibitem{Cubist2014}
Kuhn, M., Weston, S., Keefer, C., code for Cubist~by Ross~Quinlan, N.C.C.:
  Cubist: Rule- and Instance-Based Regression Modeling (2014).
\newblock {R} package version 0.0.18

\bibitem{lam2017one}
Lam, H.T., Thiebaut, J.M., Sinn, M., Chen, B., Mai, T., Alkan, O.: One button
  machine for automating feature engineering in relational databases.
\newblock arXiv preprint arXiv:1706.00327  (2017)

\bibitem{lemke2010meta}
Lemke, C., Gabrys, B.: Meta-learning for time series forecasting and forecast
  combination.
\newblock Neurocomputing \textbf{73}(10-12), 2006--2016 (2010)

\bibitem{lin2003symbolic}
Lin, J., Keogh, E., Lonardi, S., Chiu, B.: A symbolic representation of time
  series, with implications for streaming algorithms.
\newblock In: Proceedings of the 8th ACM SIGMOD workshop on Research issues in
  data mining and knowledge discovery, pp. 2--11 (2003)

\bibitem{lubba2019catch22}
Lubba, C.H., Sethi, S.S., Knaute, P., Schultz, S.R., Fulcher, B.D., Jones,
  N.S.: catch22: Canonical time-series characteristics.
\newblock Data Mining and Knowledge Discovery \textbf{33}(6), 1821--1852 (2019)

\bibitem{montero2020fforma}
Montero-Manso, P., Athanasopoulos, G., Hyndman, R.J., Talagala, T.S.: Fforma:
  Feature-based forecast model averaging.
\newblock International Journal of Forecasting \textbf{36}(1), 86--92 (2020)

\bibitem{do2019automated}
do~Nascimento~Reis, G.F.: Automated feature engineering for classification
  problems  (2019)

\bibitem{Oliveira2014EnsemblesFT}
Oliveira, M., Torgo, L.: Ensembles for time series forecasting.
\newblock In: ACML Proceedings of Asian Conference on Machine Learning. JMLR:
  Workshop and Conference Proceedings (2014)

\bibitem{paras2009feature}
Paras, S.M., Kumar, A., Chandra, M.: A feature based neural network model for
  weather forecasting.
\newblock International Journal of Computational Intelligence \textbf{4}(3),
  209--216 (2009)

\bibitem{percival2006wavelet}
Percival, D.B., Walden, A.T.: Wavelet methods for time series analysis, vol.~4.
\newblock Cambridge university press (2006)

\bibitem{Pfahringer2000}
Pfahringer, B., Giraud-Carrier, C.: Meta-learning by landmarking various
  learning algorithms.
\newblock pp. 743--750 (2000)

\bibitem{pinto2016towards}
Pinto, F., Soares, C., Mendes-Moreira, J.: Towards automatic generation of
  metafeatures.
\newblock In: Pacific-Asia Conference on Knowledge Discovery and Data Mining,
  pp. 215--226. Springer (2016)

\bibitem{prudencio2004using}
Prud{\^e}ncio, R.B., Ludermir, T.B.: Meta-learning approaches to selecting time
  series models.
\newblock Neurocomputing \textbf{61}, 121--137 (2004)

\bibitem{quinlan1993combining}
Quinlan, J.R.: Combining instance-based and model-based learning.
\newblock In: Proceedings of the tenth international conference on machine
  learning, pp. 236--243 (1993)

\bibitem{rasmussen2003gaussian}
Rasmussen, C.E.: Gaussian processes in machine learning.
\newblock In: Summer School on Machine Learning, pp. 63--71. Springer (2003)

\bibitem{robnik1997adaptation}
Robnik-{\v{S}}ikonja, M., Kononenko, I.: An adaptation of relief for attribute
  estimation in regression.
\newblock In: Machine Learning: Proceedings of the Fourteenth International
  Conference (ICML’97), vol.~5, pp. 296--304 (1997)

\bibitem{Takens1981}
Takens, F.: Dynamical Systems and Turbulence, Warwick 1980: Proceedings of a
  Symposium Held at the University of Warwick 1979/80, chap. Detecting strange
  attractors in turbulence, pp. 366--381.
\newblock Springer Berlin Heidelberg, Berlin, Heidelberg (1981)

\bibitem{taylor2018forecasting}
Taylor, S.J., Letham, B.: Forecasting at scale.
\newblock The American Statistician \textbf{72}(1), 37--45 (2018)

\bibitem{tibshirani1996regression}
Tibshirani, R.: Regression shrinkage and selection via the lasso.
\newblock Journal of the Royal Statistical Society: Series B (Methodological)
  \textbf{58}(1), 267--288 (1996)

\bibitem{wang2006characteristic}
Wang, X., Smith, K., Hyndman, R.: Characteristic-based clustering for time
  series data.
\newblock Data mining and knowledge Discovery \textbf{13}(3), 335--364 (2006)

\end{thebibliography}

\end{document}